\documentclass[letterpaper, 10 pt, conference]{ieeeconf}  

\IEEEoverridecommandlockouts

\usepackage{amsmath}
\usepackage{cite}
\usepackage{amssymb,amsfonts}
\usepackage[ruled,vlined]{algorithm2e}
\usepackage{graphicx}
\usepackage{subfloat}
\usepackage{textcomp}
\usepackage{xcolor}
\usepackage{verbatim} 
\usepackage{mathrsfs}
\usepackage{subfigure}
\usepackage{empheq}
\usepackage{courier}
\usepackage{lipsum}
\usepackage{comment}
\usepackage{gensymb}
\usepackage{graphicx}
\usepackage{mathtools, cuted}
\usepackage{bm}
\usepackage{ wasysym } 
\usepackage{soul}
\makeatletter
\let\NAT@parse\undefined
\makeatother

\usepackage[hypertexnames=true,bookmarks=false]{hyperref}
\hypersetup{
  colorlinks=false,
  pdfborderstyle={/S/S/W 1.5},
  linkbordercolor=red,
  citebordercolor=green,
  urlbordercolor=cyan
}

\usepackage{amsthm}
\newtheoremstyle{remarkcolon}
  {0.5\baselineskip} 
  {0.5\baselineskip} 
  {\itshape}         
  {0pt}              
  {\bfseries}        
  {:}                
  {0.5em}            
  {}                 
\theoremstyle{remarkcolon}
\newtheorem{remark}{Remark}

\newcommand{\vg}[1]{\bm{#1}}
\renewcommand{\v}[1]{\mathbf{#1}}

\title{\LARGE \bf
ACLM: ADMM-Based Distributed Model Predictive Control for Collaborative Loco-Manipulation
}

\author{Ziyi Zhou$^{*}$, Pengyuan Shu$^{*}$, Ruize Cao$^{*}$, Yuntian Zhao, and Ye Zhao
\thanks{*Authors have contributed equally.}
\thanks{The authors are with The Institute for Robotics and Intelligent Machines, Georgia Institute of Technology. {\tt\small \{zhouziyi, pshu8, rcao73, yzhao801, yzhao301\}@gatech.edu.}}%
}

\begin{document}

\maketitle
\thispagestyle{empty}
\pagestyle{empty}

\begin{abstract}
Collaborative transportation of heavy payloads via loco-manipulation is a challenging yet essential capability for legged robots operating in complex, unstructured environments. Centralized planning methods, e.g., holistic trajectory optimization, capture dynamic coupling among robots and payloads but scale poorly with system size, limiting real-time applicability. In contrast, hierarchical and fully decentralized approaches often neglect force and dynamic interactions, leading to conservative behavior. This study proposes an Alternating Direction Method of Multipliers (ADMM)-based distributed model predictive control framework for collaborative loco-manipulation with a team of quadruped robots with manipulators. By exploiting the payload-induced coupling structure, the global optimal control problem is decomposed into parallel individual-robot-level subproblems with consensus constraints. The distributed planner operates in a receding-horizon fashion and achieves fast convergence, requiring only a few ADMM iterations per planning cycle. A wrench-aware whole-body controller executes the planned trajectories, tracking both motion and interaction wrenches. Extensive simulations with up to four robots demonstrate scalability, real-time performance, and robustness to model uncertainty. \url{https://admm-clm.github.io/}

\end{abstract}

\section{Introduction}
\label{sec:introduction}
\begin{figure}
    \centering
    \includegraphics[width=\linewidth]{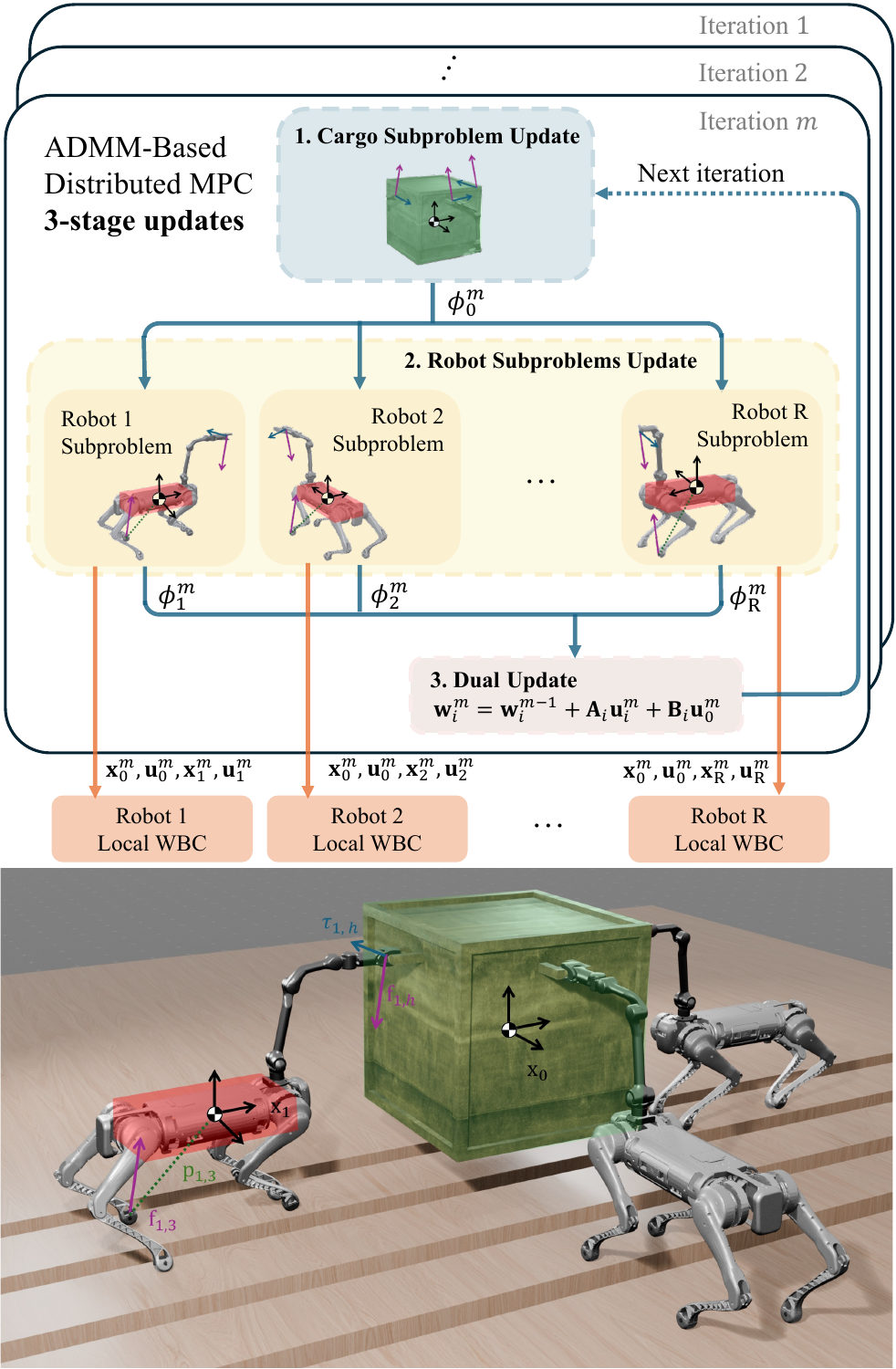}
    \vspace{-0mm}
    \caption{ADMM-based distributed MPC framework exploiting the star-shaped coupling induced by the shared payload. The robot subproblems are solved in parallel with consensus on interaction wrenches, and the resulting trajectories are executed by local WBC.}
    \label{fig:system_diagram}
    \vspace{-0mm}
\end{figure}

Recent robotics research has increasingly focused on enhancing loco-manipulation capabilities, aiming to automate repetitive and labor-intensive tasks while preserving all-terrain mobility. Among these tasks, collaborative transportation of heavy and oversized loads via loco-manipulation has attracted significant attention due to their prevalence in logistics, mining, construction, agriculture, and search-and-rescue operations \cite{farivarnejad2022multirobot}. 
Collaborative loco-manipulation, in which multiple legged robots equipped with manipulators cooperatively transport a shared load, offers a promising approach by distributing both the payload weight and the control effort across the robot team.

To simplify coordination, many approaches adopt hierarchical frameworks that decouple payload and robot planning \cite{rigo2025hierarchical,yang2022}. While scalable, these designs often rely on quasi-static assumptions or neglect dynamic and kinematic coupling between robots and the payload, leading to conservative motions. For quadrupedal manipulators, where locomotion and manipulation forces are strongly coupled, ignoring these interactions certainly degrade performance and stability especially on rough terrains (see Fig.~\ref{fig:system_diagram}), motivating planning approaches that explicitly model force and dynamic coupling.

In contrast, centralized planning frameworks can naturally account for dynamic coupling and shared constraints among the robots and the payload by solving a holistic optimal control problem (OCP) \cite{devincenti2023,sun2025agile}. However, this comes at the cost of significantly increased computational complexity. Compared to teams of wheeled mobile manipulators or quadrotors, cooperative quadrupedal manipulators present additional challenges due to a high number of degrees of freedom (DoFs), hybrid contact dynamics, and rough terrain. As a result, the scalability of fully centralized planning with respect to the number of robots becomes a critical bottleneck, limiting its applicability for real-time planning and control.

In this work, we explicitly analyze the coupling structure between quadrupedal manipulators and a shared payload and leverage the Alternating Direction Method of Multipliers (ADMM) \cite{boyd2011distributed} to decompose the resulting multi-robot optimal control problem into tractable subproblems. Although the system exhibits strong coupling through the payload dynamics, each robot interacts directly only with the payload rather than with other robot teammates. This star-shaped coupling structure enables parallel optimization across robots through carefully designed consensus constraints, allowing distributed computation while preserving the essential force and dynamic interactions induced by the shared load.

We further adopt a real-time, receding-horizon implementation of the distributed planner within a model predictive control (MPC) framework. Notably, the solution from the previous MPC window provides an effective warm start for the current optimization, allowing the ADMM solver to converge to a satisfactory consensus with only a few ADMM iterations per planning cycle. At the lower level, a wrench-aware whole-body controller (WBC) tracks the planned end-effector poses, foot placements, and interaction wrenches, bridging distributed planning and execution. Although whole-body inverse dynamics is well studied \cite{bellicoso2019alma}, to our knowledge this is the first integrated distributed MPC–WBC pipeline that accounts for coupling effects in prehensile collaborative loco-manipulation with multiple legged manipulators in challenging environments.

Our contributions can be summarized as follows:
\begin{itemize}
    \item 
    We formulate the collaborative prehensile loco-manipulation with multiple quadrupedal manipulators as a centralized OCP and develop an ADMM-based distributed MPC that decomposes it into parallel robot-level subproblems coupled through payload dynamics.
    \item By exploiting the star-shaped coupling induced by the shared payload, we treat the payload as an independent subsystem and enforce consensus only on interaction wrenches while using approximated state copies to maintain compact problem size, keeping each subproblem scalable and enabling fast online planning.
    \item We integrate the distributed planner with a wrench-aware WBC and evaluate teams of up to four robots on diverse collaborative transportation tasks, including obstacle avoidance and rough-terrain traversal, demonstrating real-time performance independent of team size (50 Hz with perceptive inputs and 100 Hz without them) and robustness to model uncertainties.
\end{itemize}

\section{Related Work}
\label{sec:related_work}
\subsection{Collaborative Transportation and Manipulation}
Multi-robot collaborative transportation has been studied on aerial and ground platforms, with approaches broadly categorized as centralized, decentralized, or leader–follower frameworks. Centralized methods scale poorly \cite{nikou2017nonlinear,sun2025agile}, while decentralized approaches trade coordination optimality for scalability \cite{khatib1996coordination,verginis2018communication,culbertson2018decentralized}. Leader–follower strategies \cite{farivarnejad2022multirobot,sugar2002control} avoid explicit group-level trajectory optimization.
In contrast, collaborative loco-manipulation with legged robots remains less explored due to high DoFs and complex dynamics.

\textbf{Holonomically constrained systems},
using cables or rigid links, enable scalable decentralized control through learning-based or hierarchical architectures \cite{pandit2024,kim2023,yang2022}, but their applicability is limited by restrictive mechanical assumptions and reduced interaction flexibility.
\textbf{Prehensile collaborative loco-manipulation} allows more flexible grasp-based interaction and has been addressed using centralized MPC with simplified models \cite{devincenti2023}, hierarchical planning frameworks \cite{rigo2025hierarchical}, passive-arm leader–follower designs \cite{turrisi2024}, and learning-based decentralized strategies \cite{pandit2025multi,an2025collaborative}. However, many of these approaches depend on centralized optimization that limits scalability, requires strong hierarchy, or need complex reward design.
\textbf{Non-prehensile approaches} rely on indirect manipulation such as pushing or carrying \cite{feng2025learning,sombolestan2024,ji2021reinforcement}, typically within hierarchical planners. While suitable for long-horizon tasks, they lack direct force control and manipulation precision.

Motivated by the flexibility of prehensile manipulation and the scalability limitations of existing centralized approaches, we focus on prehensile collaborative loco-manipulation and propose a distributed MPC framework based on distributed optimization, which enables scalable subsystem updates with local communication while retaining much of the centralized solution quality and improving computational scalability.

\subsection{ADMM for Multi-Robot Distributed Control}
Many OCPs exhibit intrinsic distributed structure despite coupling effects \cite{zhao2024survey}. In multi-robot systems, spatial separability enables decomposition via distributed optimization, particularly ADMM \cite{boyd2011distributed}. Multi-robot path finding (MAPF) is an example where these ideas have been successfully applied \cite{saravanos2023distributed,tajbakhsh2025asynchronous}: robots maintain local copies of their own and neighboring agents' trajectories, enforce consensus over coupling constraints such as collision avoidance, and achieve scalability to large teams. However, these formulations mostly consider simplified robot models, which keep subproblem computationally tractable despite the augmented state dimension for each robot.

Collaborative manipulation also exhibits an implicit distributed structure, although the subsystems are coupled through shared object dynamics. 
To avoid the growing number of control inputs for the robot-object subsystem, the authors in \cite{shorinwa2020scalable} propose a decomposition in which each subsystem optimizes only the force and torque contributed by a single robot, while consensus is enforced on the object trajectory. A similar idea is extended to contact-implicit manipulation settings in \cite{shorinwa_disco_2024}. 
In contrast, collaborative loco-manipulation with legged manipulators involves high-dimensional, multi-contact dynamics. Rather than augmenting robot states with object states, we treat the object as an independent subsystem and couple subsystems only through interaction forces and torques under consensus. This keeps subproblems compact and enables scalable distributed MPC despite nonlinear, high-dimensional dynamics.

\section{Preliminaries}
\label{sec:preliminaries}
\subsection{Consensus ADMM}
We adopt the standard \textit{consensus} ADMM formulation \cite{boyd2011distributed} for problems of the form
\begin{equation}\label{eq:consensusADMM}
\begin{aligned}
    \underset{\Bar{\v x},\{\v x_i\}_{i=1}^{P}}{\text{min}} \quad
    & \sum_{i = 1}^{P} f_i(\v x_i) + g(\Bar{\v x}) \\
    \text{s.t.} \quad
    & \v x_i = \Bar{\v x}, \quad i = 1,\dots,P ,
\end{aligned}
\end{equation}
where $\mathbf{x}_i$ denotes the local decision variables of subsystem $i$ among $P$ subsystems, $\Bar{\v x}$ is a global consensus variable, and $g(\Bar{\v x})$ is an optional regularization term. 
By establishing the scaled augmented Lagrangian (AL) of~\eqref{eq:consensusADMM},
consensus ADMM proceeds by alternating updates of the local variables, the consensus variable, and the dual variables $\{\v w_i\}$:
\begin{subequations}\label{eq:consensus-updates}
\begin{align}
    \v x_i^{m+1}
    &:= \arg\min_{\v x_i}
    \left(
        f_i(\v x_i)
        + \frac{\rho}{2}
        \|\v x_i - \Bar{\v x}^m + \v w_i^m\|^2
    \right),
    \quad \forall i, \label{eq:x-update} \\[4pt]
    \Bar{\v x}^{m+1}
    &:= \arg\min_{\Bar{\v x}}
    \left(
        g(\Bar{\v x})
        + \frac{\rho}{2}
        \sum_{i=1}^{N}
        \|\v x_i^{m+1} - \Bar{\v x} + \v w_i^m\|^2
    \right), \label{eq:z-update} \\[4pt]
    \v w_i^{m+1}
    &:= \v w_i^m + \v x_i^{m+1} - \Bar{\v x}^{m+1},
    \quad \forall i. \label{eq:dual-update}
\end{align}
\end{subequations}
where $m$ is the ADMM iteration number and $\rho > 0$ is the penalty parameter.
This structure enables parallel optimization of the local subproblems~\eqref{eq:x-update}, followed by a lightweight consensus update~\eqref{eq:z-update}. The updates in~\eqref{eq:consensus-updates} correspond to standard \textit{consensus} ADMM. 
We use the standard \textit{Gauss–Seidel} consensus variant, which provides better convergence than the fully parallel \textit{Jacobi} variant while keeping the consensus subproblem inexpensive.

\subsection{Centralized Optimization Formulation}
The centralized optimization formulation for collaborative loco-manipulation is written as:
\begin{subequations} 
\begin{align}
	\underset{\v U, \v X}{\text{min}} \ \mathcal{C}(\v U, \v X) \coloneqq & \sum_{k = 0}^{N-1}l_k(\v x[k], \v u[k]) + l_N(\v x[N]) \\
	\text{s.t.} \quad & \v x[k+1] = \mathcal{D}(\v x[k], \v u[k]) \\
    & \v x[0] = \v x_{\text{init}} \\
    & \v g(\v x[k], \v u[k]) = 0 \\
    & \v h(\v x[k], \v u[k]) \leq 0
\end{align}
\end{subequations}
where $\v x[k]$ and $\v u[k]$ denote the state and control variables at the $k$-th time step over a horizon of $N+1$ steps, with the initial condition $\v x_{\text{init}}$. The stacked state and control trajectories are defined as $\v X := [\v x[0]^{\top}, \v x[1]^{\top}, \dots,\v x[N]^{\top}]^{\top}$ and $\v U := [\v u[0]^{\top}, \v u[1]^{\top}, \dots,\v u[N-1]^{\top}]^{\top}$. At each time step, the state and control vectors are decomposed into payload and robot components: $\v x[k] := [\v x_{0}[k]^{\top}, \dots, \v x_{i}[k]^{\top}, \dots,\v x_{R}[k]^{\top}]^{\top}$, $\v u[k] := [\v u_{1}[k]^{\top}, \dots, \v u_{i}[k]^{\top}, \dots,\v u_{R}[k]^{\top}]^{\top}$,
where index $i \in \{0,1,\dots,R\}$ denotes the $i$-th subsystem, with $i=0$ corresponding to the payload and $i>0$ to the robots. The payload has no direct control input, as its motion is governed by the interaction forces and torques exerted by the robots. The payload state is defined as
\begin{equation}
    \begin{aligned}
        \v x_{0}[k] := [\v r_{0}[k]^{\top}, \dot{\v r}_{0}[k]^{\top}, \vg \theta_{0}[k]^{\top}, \v l_{0}[k]^{\top}]^{\top}
    \end{aligned}
\end{equation}
where $\v r$, $\dot{\v r}$, $\vg \theta$, and $\v l \in \mathbb{R}^3$ denote the position, linear velocity, Euler angles, and angular momentum of the rigid body, respectively. In this work, each robot is approximated as a single rigid body to reduce optimization complexity; however, the proposed distributed framework can readily extend to full-order models with joint-level dynamics and whole-body constraints. For each robot $i \in \{1,\dots,R\}$, the state and control variables are defined as
\begin{equation}
    \begin{aligned}
        & \v x_{i}[k] := [\v r_{i}[k]^{\top}, \dot{\v r}_{i}[k]^{\top}, \vg \theta_{i}[k]^{\top}, \v l_{i}[k]^{\top}, \v p_{i}[k]^{\top}]^{\top}\\
        & \v u_{i}[k] := [\v f_{i}[k]^{\top}, \dot{\v p}_{i}[k]^{\top}, \v f_{i,h}[k]^{\top}, \vg \tau_{i,h}[k]^{\top}]^{\top}
    \end{aligned}
\end{equation}
where $\mathbf{p}_i[k] := \big[\mathbf{p}_{i,0}[k]^\top,\dots,\mathbf{p}_{i,n_f-1}[k]^\top\big]^\top$ and $\mathbf{f}_i[k] := \big[\mathbf{f}_{i,0}[k]^\top,\dots,\mathbf{f}_{i,n_f-1}[k]^\top\big]^\top$ 
stack the positions ($\dot{\v p}_i$ as velocities) and contact forces of all $n_f$ feet, 
with $j \in \{0,\dots,n_f-1\}$ indexing the feet. 
Here, $\mathbf{f}_{i,h}$ and $\boldsymbol{\tau}_{i,h}$ denote the manipulation force and torque applied at the arm end-effector (EE).

The coupled system dynamics $\mathcal{D}$ can be decomposed into payload and robot components $\mathcal{D}^0$ and $\mathcal{D}^{i}$. For clarity of presentation, the time-step superscript $k$ is omitted in the following equations. The second-order dynamics of the payload are given by
\begin{subequations}\label{eq:raw_payload_dynamics}
    \begin{align}
        \ddot{\v r}_{0} &= \frac{1}{m_0}\!\left(- \sum\nolimits_i \v f_{i,h}\right) + \boldsymbol{g}, \\
        \dot{\v l}_0 &= \sum\nolimits_i 
        \big(\v p_{i,h}(\v r_0,\vg \theta_0) - \v r_0\big) \times (-\v f_{i,h})
        - \vg \tau_{i,h},
    \end{align}
\end{subequations}
where $\v p_{i,h}(\cdot,\cdot)$ denotes the position of the $i$-th robot’s arm EE expressed in the inertial frame, obtained by transforming a constant handle offset in the cargo frame. The second-order dynamics of the $i$-th robot are expressed as
\begin{subequations}\label{eq:raw_robot_dynamics}
    \begin{align}
        \ddot{\v r}_{i} =&
        \frac{1}{m_i}\!\left(\sum\nolimits_{j} \v f_{i,j} + \v f_{i,h}\right) + \boldsymbol{g}, \\
        \dot{\v l}_i =&
        \sum\nolimits_j (\v p_{i,j} - \v r_i) \times \v f_{i,j}
        + \\ & (\v p_{i,h}(\v r_0,\vg \theta_0) - \v r_i) \times \v f_{i,h}
        + \vg \tau_{i,h},
    \end{align}
\end{subequations}

All continuous-time dynamics are discretized using the backward Euler method for use in the optimal control formulation. The equality and inequality constraint functions $\v g(\cdot)$ and $\v h(\cdot)$ are introduced later.

\section{Distributed Optimization}
\label{sec:approach}

In the previous section, it can be seen that each dynamics equation and part of the inequality constraints still depend not only single payload/robot state but also other system's states. For the latter case, we call them inter-system constraints. Therefore, one major contribution of this work is to develop an optimization framework that further separates each subsystem so that each subproblem can be solved in parallel to improve the computational efficiency.

\subsection{Decomposed System Dynamics}
To apply ADMM, we first create the control variables $\v u_0$ for the payload, which consists of global copies of the manipulation wrench, including force and torque variables applied by each robot:
\begin{equation}
    \begin{aligned}
        \v u_{0}[k] := [\bar{\v f}_{i,h}[k]^{\top}, \bar{\vg \tau}_{i,h}[k]^{\top}]^{\top}, i \in \{1, \dots, R\}
    \end{aligned}
\end{equation}

For better illustration of each subsystem, we use $\vg \phi_i = [\v x_i[0,\dots,N], \v u_i[0,\dots,N-1]]^{\top}$ to express the trajectory of the concatenated state and control variables for the $i$-th rigid body. Then we build consensus constraint for the manipulation force and torque variables between the payload and robot states based on the Newton's law:
\begin{subequations}\label{eq:consensus_constraint}
    \begin{align}
        \nonumber & \forall i \in \{1, \dots, R\},\\
        & \v f_{i,h}[k] + \bar{\v f}_{i,h}[k] = 0,\\
        & \vg \tau_{i,h}[k] + \bar{\vg \tau}_{i,h}[k] = 0
    \end{align}
\end{subequations}

\begin{remark}\label{remark:consensus}
\textbf{Approximated State Copy}: Instead of introducing consensus copies of both state and control variables for each robot’s neighbor (the payload), we deliberately avoid enforcing state consensus to prevent unnecessary state augmentation. For a single quadrupedal manipulator, even using a simplified model, the subsystem already has a high-dimensional state (i.e., 24 states), making the computational cost highly sensitive to further state expansion. Moreover, unlike contact-implicit formulations \cite{shorinwa_disco_2024}, the dynamic coupling between the robots and the payload in our static grasping setup is dominated by interaction forces and torques rather than state variables. Consequently, for dynamics and inter-system constraints that involve state variables from other subsystems, we use the values from the previous ADMM iteration. This approximation has been empirically shown to have negligible impact on performance while improving computational efficiency, consistent with observations in \cite{amatucci2022}.
\end{remark}
By establishing local copies, the payload dynamics gets rid of the dependency on $\v u_i (i > 0)$ and replaces it with the new control variable $\v u_0$. To consider the coupling effects from the payload state in the robot dynamics without creating payload state copies, we use the payload states from the previous ADMM iteration as described in Remark \ref{remark:consensus}. The discrete dynamics equations for the payload and robots at the $m+1^{\text{th}}$ ADMM iteration can be written as:
\begin{subequations}
    \begin{align}
        & \v x_{0}^{m+1}[k+1] = \mathcal{D}^0(\v x_{0}^{m+1}[k], \v u_{0}^{m+1}[k])\\
        & \v x_{i}^{m+1}[k+1] = \mathcal{D}^i(\v x_{i}^{m+1}[k], \v x_{0}^{m}[k], \v u_{i}^{m+1}[k])
    \end{align}
\end{subequations}
Then at the $m+1^{\text{th}}$ ADMM iteration, the states from $m^{\text{th}}$ ADMM iteration can be treated as constant variables. 

\subsection{Constraints}
We classify the constraints into independent constraints, applied locally to each subsystem, and inter-system coupling constraints, handled through ADMM.

\subsubsection{Independent --- \textbf{Frictional and Contact}}
The manipulation contact wrench cone is simplified to a positive normal force constraint, a manipulation friction cone, and a bounding box $\mathcal{B}$ for the moment part defined in Eqs.~(\ref{eq:mani_force_non_zero}) - (\ref{eq:wrench_cone}).
Point-foot positive normal force and frictional constraints are defined in Eqs.~(\ref{eq:force_non_zero}) - (\ref{eq:force_friction_cone}). We define $\v n$ as the normal vector of the grasping point or the ground, corresponding to the payload orientation $\vg \theta_0$ or the foot EE tangential position $\v p^{xy}_{i,j}$ by inquiring the terrain map. The friction cone $\mathcal{F}$ depends on the friction coefficient $\mu$ and the $\v n$. Contact constraints force the EE velocity to be zero during stance phase and the contact force to be zero during non-contact. 
\begin{subequations}
\begin{align}
    \nonumber &\forall i \in \{1,\dots,R\},\\\label{eq:mani_force_non_zero}
    & \v f_{i,h}[k] \cdot \v n_{i,h}(\vg \theta_0[k]) \ge 0,\\
    \label{eq:wrench_cone}
        & \v f_{i,h}[k] \in \mathcal{F}(\mu, \v n_{i,h}), \vg \tau_{i,h} \in \mathcal{B}_{i,h}
        \\\label{eq:force_non_zero}
    & \v f_{i,j}[k] \cdot \v n_{i,j}(\v p_{i,j}^{xy}[k]) \ge 0,
    \ \forall j \in [n_f]
    \\\label{eq:force_friction_cone}
        & \v f_{i,j}[k] \in \mathcal{F}(\mu, \v n_{i,j}), 
        \ \forall j \in [n_f]
\end{align}
\end{subequations}


\subsubsection{Independent --- \textbf{Foot Kinematics}}
The kinematics constraint limits the possible foot EE movements inside a nominal box region to ensure safety and avoid singularity.

\subsubsection{Independent --- \textbf{Foot Placement}}


Foot placements are constrained to satisfy either a default terrain height or the elevation map. In perceptive mode, admissible footholds are restricted to convex safe regions extracted from the elevation map, ensuring collision-free and stable contacts. The convex regions are selected heuristically for each stance phase similar to \cite{grandia2023perceptive}.


\subsubsection{Independent --- \textbf{Collision Avoidance}}
Obstacle avoidance is incorporated via a control barrier function (CBF) formulation, maintaining forward invariance of the safe set for both robots and payload.
Similarly, a terrain-aware constraint is imposed on the payload to prevent collisions during terrain transitions, such as moving from flat ground to a slope.

\subsubsection{Inter-System --- \textbf{Arm Kinematics}}
The arm kinematics constraint bounds the relative pose between each robot base and the grasping handle (arm EE), ensuring the grasp remains within a feasible workspace. Therefore, similar to the previous method for decomposing the system dynamics, we use the state from the previous ADMM iteration. Note that due to the fixed state assumption, we add additional constraints for the payload with fixed robot states, maintaining the constraint feasibility from both payload and robots. 
\begin{equation}
\begin{aligned}
    & \v r_0 [k], \vg \theta_0[k] \in \mathcal{R}_h(\v r_i^{m}[k],\vg \theta_{i}^{m}[k], \v p_h^{\text{max}})\\ 
    \nonumber
    & \forall i \in \{1,\dots,R\}, j \in [n_f]\\
    \label{eq:arm_rom}
    & \v r_{i}[k], \vg \theta_{i}[k] \in \mathcal{R}_h(\v r_0^{m}[k],\vg \theta_{0}^{m}[k], \v p_h^{\text{max}})
\end{aligned}
\end{equation}
where $\mathcal{R}_h$ is defined as a 3D box constraint around a nominal arm workspace based on the anchored system's position and orientation, and constrained by a maximum deviation $\v p_h^{\text{max}}$.

\subsubsection{Inter-System --- \textbf{Formation}}
The formation constraint keeps each robot’s center of mass near its nominal offset from the payload in the payload frame. Although similar to the arm kinematics constraint, it empirically enhances team-level obstacle avoidance and mitigates the risk of individual robots or the payload getting trapped in local minima. As with other inter-system constraints, it is enforced in a distributed manner by using the other subsystem’s state from the previous ADMM iteration.


\subsection{Cost Function}
The cost function is directly defined for each subsystem separately, consisting of simple tracking costs and regularization terms governed by diagonal matrices $\v Q$ and $\v R$. 
\begin{equation}
    l_k^{i} = \delta \vg \phi_Q[k]^T\mathbf{Q}\hspace{2pt}\delta \vg \phi_Q[k] + \vg \phi_R[k]^{T}\mathbf{R}\vg \phi_R[k]
\end{equation}
where $\delta \vg \phi_Q = \v x_i - \v x_i^{\text{ref}}$ and $\vg \phi_R = \v u_i$.
Tracking costs include deviation from the desired base reference trajectories. The regularization term includes minimizing the foot and arm EE velocities, and contact forces and wrenches to encourage motion smoothness. 

\subsection{Distributed Update}
Now we are able to adopt \textit{consensus} ADMM to update each subproblem separately. Based on the decomposition from previous sections, the overall optimization for each ADMM iteration now can be further written as:
\begin{subequations} 
\begin{align}
	\underset{\vg \phi_0,\dots,\vg \phi_N}{\text{min}} \ \mathcal{C} \coloneqq & \sum_{i = 0}^{R}(\sum_{k = 0}^{N-1}l_k^i(\v x_{i}[k], \v u_{i}[k]) + l_N^i(\v x_{i}[N])) \\\label{eq:payload_dynamics}
	\text{s.t.} \quad & \v x_{0}[k+1] = \mathcal{D}^0(\v x_{0}[k], \v u_{0}[k]) \\\label{eq:payload_constraint}
    & \v h_0(\v x_{0}[k], \v u_{0}[k]) \leq 0\\
    & \forall i \in \{1, \dots, R\} \nonumber\\\label{eq:robot_dynamics}
    & \v x_{i}[k+1] = \mathcal{D}^i(\v x_{i}[k], \v u_{i}[k]) \\\label{eq:robot_eq_constraint}
    & \v g_i(\v x_{i}[k], \v u_{i}[k]) = 0 \\
    \label{eq:robot_ineq_constraint}
    & \v h_i(\v x_{i}[k], \v u_{i}[k]) \leq 0 \\
    \label{eq:consensus_constraint_vecterized}
    & \v A_i \v u_i + \v B_i \v u_0 = 0
\end{align}
\end{subequations}
where $\v A_i$ and $\v B_i$ denote the vectorized forms of the consensus constraints defined in Eqs.~(\ref{eq:consensus_constraint}), which constitute the only coupling constraints among the subproblems.

Then, based on ADMM, the original problem is divided into $N+1$ subproblems which are also known as sub-blocks. Each sub-block only requires part of the aforementioned AL as the local cost function with the payload sub-block being:
\begin{equation}
    \begin{aligned}
        \mathcal{L}_0(\cdot) = & \sum_{k = 0}^{N-1}l_k^0(\v x_{0}[k], \v u_{0}[k]) + l_N^0(\v x_{0}[N]) + \\
        & \sum_{i = 1}^{R} \frac{\rho}{2}\|\v A_i \v u_i+\v B_i \v u_0 + \v w_i\|^2
    \end{aligned}
\end{equation}

The $i^{\text{th}}$ robot sub-block can be formulated as:
\begin{equation}
    \begin{aligned}
        \mathcal{L}_i(\cdot) = & \sum_{k = 0}^{N-1}l_k^i(\v x_{i}[k], \v u_{i}[k]) + l_N^i(\v x_{i}[N]) + \\
        & \frac{\rho}{2}\|\v A_i \v u_i+\v B_i \v u_0 + \v w_i\|^2
    \end{aligned}
\end{equation}
Then as shown in Fig.~\ref{fig:system_diagram}, for each ADMM iteration $m$, the updating sequence in a scaled form is
\begin{subequations}\label{eq:updates}
    \begin{align}
       \label{eq:primal-a}
        &\vg \phi_0^{m+1}=\underset{\vg \phi_0}{\arg\min} \ \mathcal{L}_{0}(\vg \phi_0, \vg \phi_1^{m},\dots, \vg \phi_N^{m}, \v w^{m})  \\\nonumber &\hspace{1.2cm} \text{s.t. Eq. }(\rm \ref{eq:payload_dynamics}), \text{Eq. }{\rm (\ref{eq:payload_constraint})}
        \\ 
        \label{eq:primal-b-i} 
        &\vg \phi_i^{m+1}=\underset{\vg \phi_i}{\arg\min} \ \mathcal{L}_{i}(\vg \phi_0^{m+1}, \vg \phi_i, \v w^{m})\\
        \nonumber &\hspace{1.2cm} \text{s.t. Eq. }(\rm \ref{eq:robot_dynamics}), \text{Eqs. }{\rm (\ref{eq:robot_eq_constraint} - \ref{eq:robot_ineq_constraint})}\\
        &\v w_{i}^{m+1}=\v w_{i}^{m}+\v A_i \v u_i^{m+1}+\v B_i \v u_0^{m+1}
    \end{align}
\end{subequations}
where for each subproblem, due to the nonlinear dynamics, we employ a Riccati-like Sequential Quadratic Programming (SQP) solver to efficiently get the reduced-size subproblem solutions. All inequality constraints are penalized with a relaxed log barrier. 
Lastly, based on the consensus constraints defined in (\ref{eq:consensus_constraint}), we define the vectorized residual as $\v s = [\v A_1 \v u_1 + \v B_1 \v u_0, \dots, \v A_R \v u_R + \v B_R \v u_0]^{\top}$. Similar to \cite{zhou2020accelerated}, the stopping criteria is based on a time-scaled sum squared error with a tolerance $\epsilon$: $dt \cdot \sum_k \|\v s[k]\|_2 < \epsilon$.

\begin{figure*}[htb]
    \centering
    \includegraphics[width=\linewidth]{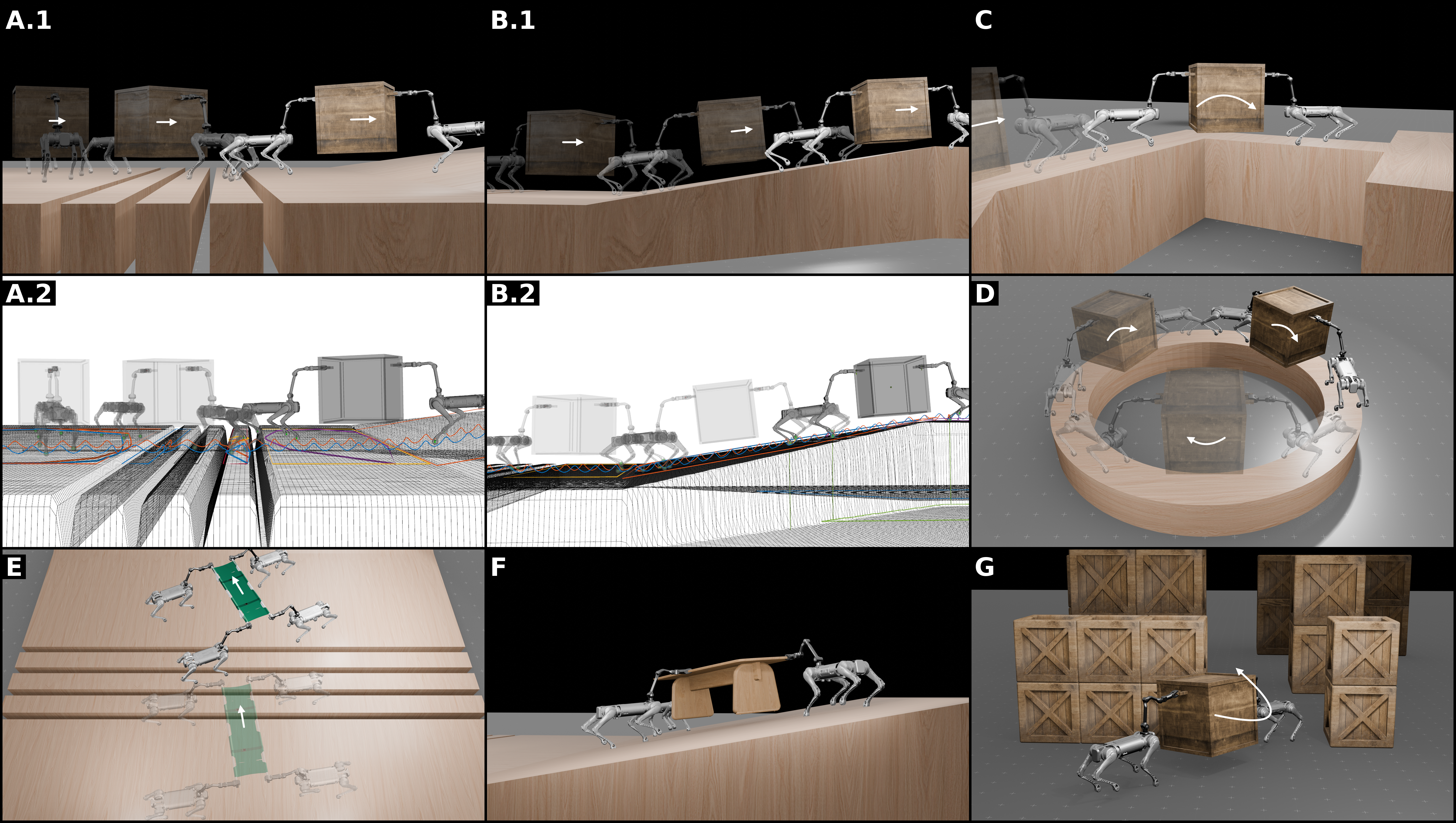}
    \vspace{-3mm}
    \caption{Illustration of diverse terrain navigation scenarios used to evaluate collaborative loco-manipulation. \textbf{A.1--A.2}: Two robots carry a cargo across stepped terrain of uniform surface height, shown in rendered and RViz views. \textbf{B.1--B.2}: Two robots carry a cargo on a sloped surface, shown in rendered and RViz views. \textbf{C}: Two robots carry a cargo through a 90-degree turn in a narrow passage. \textbf{D}: Two robots carry a cargo following a circular path on an elevated annular platform. \textbf{E}: Four robots carry a folding stretcher across stepped terrain of uniform surface height. \textbf{F}: Three robots carry a wooden table on a sloped surface. \textbf{G}: Two robots carry a cargo performing obstacle avoidance on flat terrain.}
    \label{fig:perceptive_terrain_scenarios}
    \vspace{-3mm}
\end{figure*}

\section{Real-Time Planning and Control}
\subsection{Warm-Start for Distributed MPC}
The distributed MPC framework involves iterative replanning based on Eq.~(\ref{eq:updates}). To accelerate convergence, we employ two warm-start strategies. First, within a single MPC solve, the solution of each subproblem from the previous ADMM iteration is used to initialize the corresponding subproblem in the current iteration. Second, across successive MPC solves, the primal and dual solutions from the previous window are used to initialize the first ADMM iteration of the new window. For the initial time window, the dual variables are initialized to zero, and the reference trajectory is used as the primal initial guess.

\subsection{Wrench-Aware Whole-Body Control}

The wrench-aware WBC executes the planned motion and interaction wrenches at 500 Hz.
It tracks the optimized base, foot, and arm EE poses together with the desired manipulation wrench, solving for generalized accelerations, contact forces, and joint torques. Inverse kinematics (IK) provides joint position and velocity references for both legs and arm, and the final torque command combines WBC feedforward torques with low-gain PD feedback to realize the desired motion and interaction wrenches while compensating residual tracking errors.

A central design choice is the task hierarchy. We adopt a hierarchical quadratic program (QP) \cite{bellicoso2019alma} with three priority levels, so that safety and dynamic consistency are never traded off for tracking. In the highest priority, we enforce: the floating-base equations of motion, joint torque limits, friction cones for feet and arm contact, no-contact motion for swing legs, and the arm contact wrench tracking task given the MPC desired states. In the second level, we handle base pose, swing-leg task, and arm EE pose tracking, while in the lowest level, we track the leg contact force and regularize torque/joint-acceleration. Unlike \cite{bellicoso2019alma}, we elevate wrench tracking to the highest priority in our dynamic manipulation task, thereby guaranteeing strict wrench consistency while exploiting the remaining redundancy for pose objectives.



\section{Results}
\label{sec:experiment}
\subsection{Experimental Setup}
Unless otherwise specified, we use at most 2 ADMM iterations and 1 SQP iteration per MPC solve. The rationale is discussed in Sec.~\ref{subsec:scalability_convergence}. We refer to different iteration limits as ADMM-SQP iterations. The consensus constraint tolerance is set to $5 \times 10^{-3}$. All experiments are conducted on Unitree B1-Z1 via an Intel Core i7-14650HX CPU. The implementation is majorly built upon OCS2 \cite{OCS2}. The perception modules consist of elevation mapping and convex segmentation implemented in \cite{miki2022elevation}.

We evaluate two aspects of the proposed framework: 1) \textbf{Scalability} and \textbf{Convergence}: Experiments are conducted in a non-physical simulator similar to the one used in \cite{devincenti2023} by rolling out the system dynamics in (\ref{eq:raw_payload_dynamics}) and (\ref{eq:raw_robot_dynamics}) using only the first-step MPC optimized inputs. The full-body motions are realized through IK without WBC, allowing us to isolate trajectory quality and MPC solving time from tracking errors or lower-level controller effects. 2) \textbf{Tracking performance}: Experiments are performed in a high-fidelity physical simulator with Gazebo physics engine and rendered in Blender. The MPC and WBC run asynchronously to emulate realistic multi-rate interactions between planning and control. The payloads consist of a wide range of geometries and mass including cargo box, table, and folding stretcher.


\subsection{Rough Terrain}
Fig.~\ref{fig:perceptive_terrain_scenarios} summarizes results across rough terrain scenarios. In the Gap case (A.1–A.2), the two-robot team clears steps terrain with adaptive swing heights while maintaining stable payload transport. In the Slope case (B.1–B.2), the system traverses a $10^\circ$ incline with robot base and cargo orientation compensating for terrain tilt. The Narrow Turn scenario (C) presents the most constrained configuration, requiring coordinated replanning as the team redirects the payload through a $90 \degree$ turn within a narrow passage. The annular platform scenario (D) demonstrates stable formation tracking under continuous curvature on an elevated circular path.

\subsection{Scalability and Convergence Analysis}\label{subsec:scalability_convergence}
\begin{figure}
    \centering
    \includegraphics[width=\linewidth]{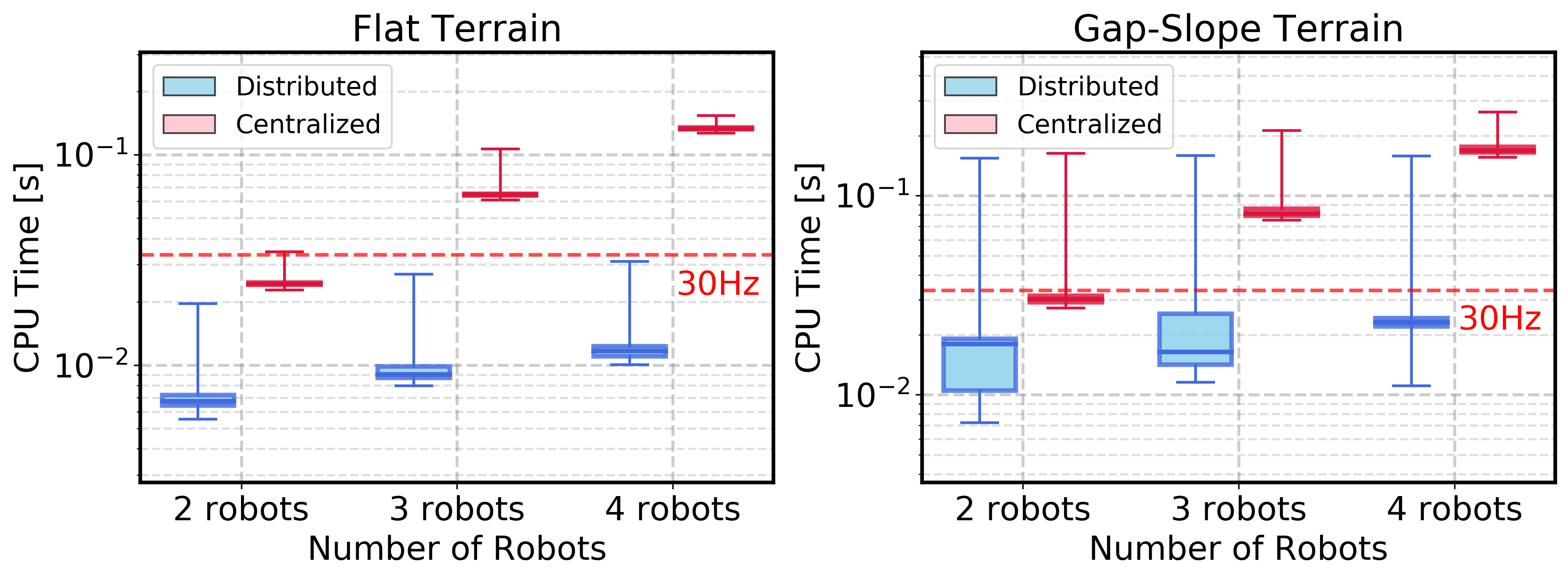}
    \vspace{-6.0mm}
    \caption{CPU time scalability comparison between distributed and centralized MPC for 2–4 robots in flat and Gap-Slope terrain scenarios. 
    }
    \label{fig:scalability_analysis}
    \vspace{-4.0mm}
\end{figure}

\begin{figure}
    \centering
    \includegraphics[width=0.95\linewidth]{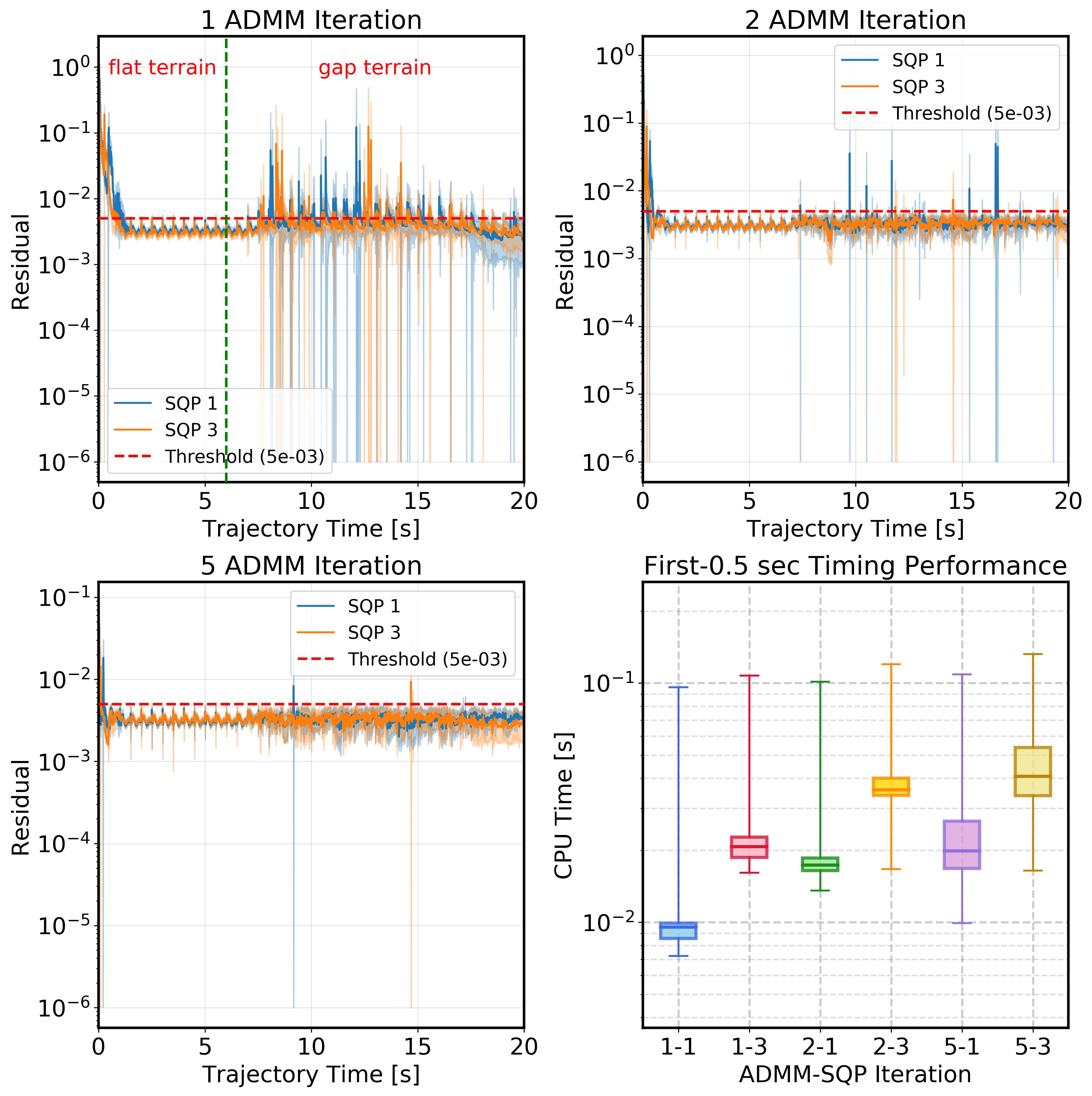}
    \vspace{-2.0mm}
    \caption{Residual convergence (top-left, top-right, and bottom-left) and computation time (bottom-right) for different ADMM-SQP iterations.}
    \label{fig:convergence_analysis}
    \vspace{-4.0mm}
\end{figure}

We use the flat and Gap-Slope (Fig.~\ref{fig:perceptive_terrain_scenarios}-E and F) terrain scenarios as benchmarks for scalability tests between distributed and centralized MPC. Fig. \ref{fig:scalability_analysis} shows computational advantages for distributed MPC that increase with robot team size. In flat terrain, distributed MPC achieves $3.6 \times$, $7.1 \times$, and $11.4 \times$ speedups for 2, 3, and 4 robots, with median CPU times of $6.73$ ms to $11.63$ ms compared to centralized MPC's $24.38$ ms to $133.13$ ms. In Gap-Slope terrain with perceptive constraints, distributed MPC achieves $1.7 \times$, $5.0 \times$, and $7.3 \times$ speedups with median times of $18.01$ ms, $16.39$ ms, and $23.03$ ms.
Distributed MPC remains 50 Hz (100 Hz on flat terrain) for all team sizes, whereas centralized MPC is below 30 Hz for 3 and 4 robots. 
The largest runtime outliers occur during the initial solve due to perception initialization and cold starts.
The superior scalability of distributed MPC stems from decomposing the multi-robot optimization into smaller, parallelizable subproblems, resulting in nearly uniform computational time across team sizes, 
whereas centralized MPC must solve a single, linearly growing problem that becomes computationally prohibitive as team size increases.

To evaluate ADMM convergence and warm-start effects, we run 60 trials on a Gap terrain (Fig.~\ref{fig:perceptive_terrain_scenarios}-A) with different ADMM-SQP iterations. Each trial uses a waypoint sent across the gap with random offsets of $\pm 1$ m in x-y and $\pm 90\degree$ in yaw. As shown in Fig. \ref{fig:convergence_analysis}, with 1 ADMM iteration, the residual drops more slowly initially along with trajectory time than with more ADMM iterations, and deviation increases, especially in the gap phase, where the consensus residual becomes unsuppressed. More SQP iterations slightly reduce large constraint violations. With 2 ADMM iterations, the residual stays within the threshold with rare violations; 5 iterations improve further. To capture computational differences, we record computation time for the first 0.5 s of planned trajectory after MPC starts, since the initial steps must quickly reduce the residual for consensus and the solver starts from scratch. 
Under the same SQP iteration, more ADMM iterations converge faster initially but are more expensive and show sparser time distributions, as more ADMM iterations are needed to converge, slowing the process initially. More SQP iterations substantially increase computation time under the same ADMM iteration. Therefore, we chose 2 ADMM iterations and 1 SQP iteration to balance constraint satisfaction and computational efficiency.

\subsection{Obstacle Avoidance}

\begin{figure}
    \centering
    \includegraphics[width=\linewidth]{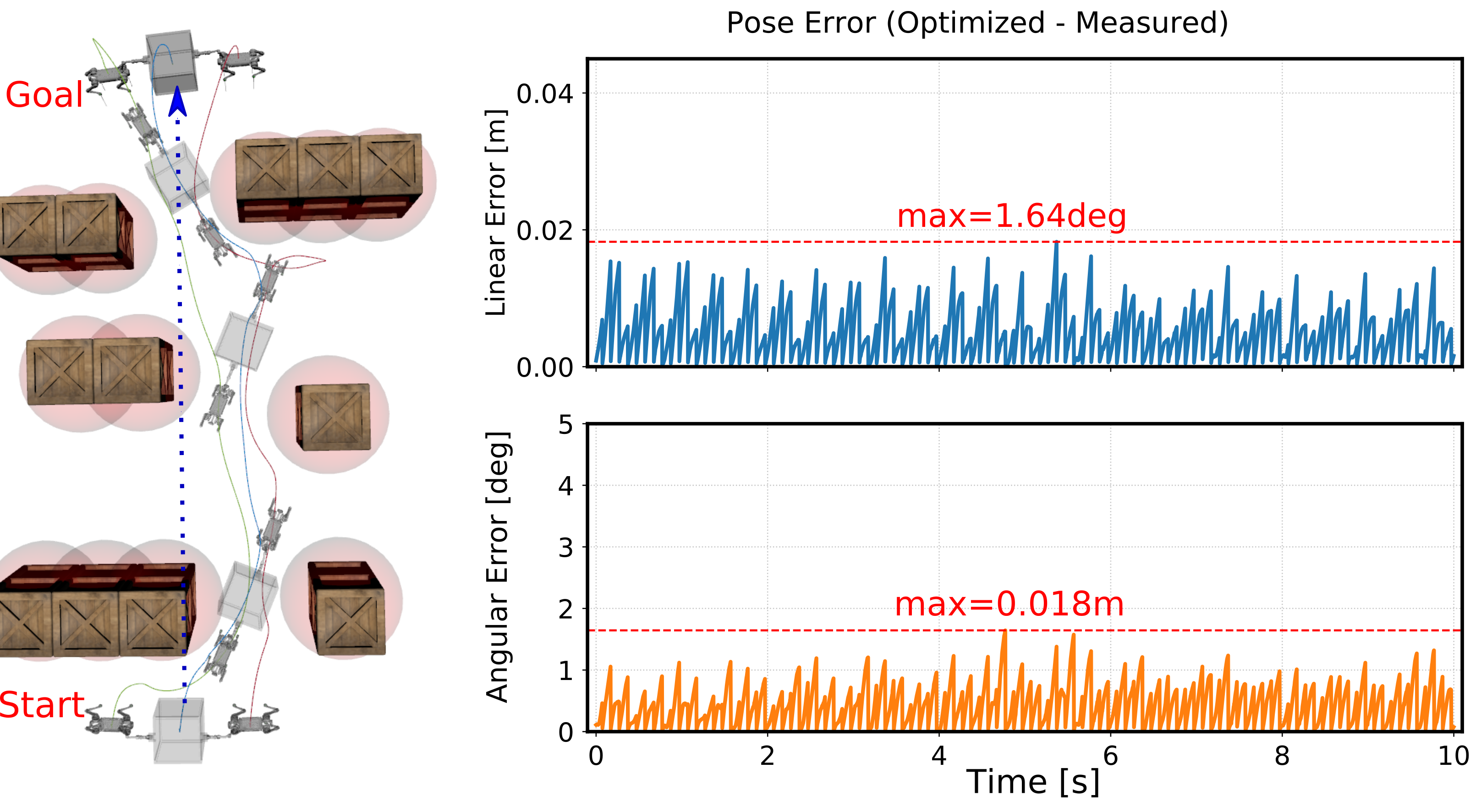}
    \vspace{-8.0mm}
    \caption{Obstacle avoidance and pose tracking performance. 
    }
    \label{fig:obstacle_avoidance_overview}
\end{figure}

To evaluate the obstacle avoidance capability of the proposed MPC framework, a scenario is constructed in a physical simulator as shown in Fig.~\ref{fig:perceptive_terrain_scenarios}-G and Fig.~\ref{fig:obstacle_avoidance_overview}. 
The left portion of Fig.~\ref{fig:obstacle_avoidance_overview} illustrates the resulting trajectory, showing that the robot--cargo system safely traverses among all obstacles. 
With only a simple start-to-goal interpolation as the initial reference, the MPC automatically finds a feasible trajectory that turns and leverages manipulator flexibility to pass narrow passages while remaining close to the reference. 
The right portion of Fig. \ref{fig:obstacle_avoidance_overview} shows the pose tracking error over a representative $10$~s segment of the full $110$~s duration, obtained from integrated MPC-WBC. 
The maximum linear error is $0.018$~m, and the maximum angular error is $1.64^\circ$. Periodic spikes in the error correspond to ground impact events during trotting, which introduce transient disturbances but remain bounded. These results demonstrate that the proposed MPC--WBC framework achieves reliable real-time obstacle avoidance while maintaining accurate pose tracking.

\subsection{Robustness and Ablation Study}
\begin{figure}[h]
    \centering
    \includegraphics[width=0.9\linewidth]{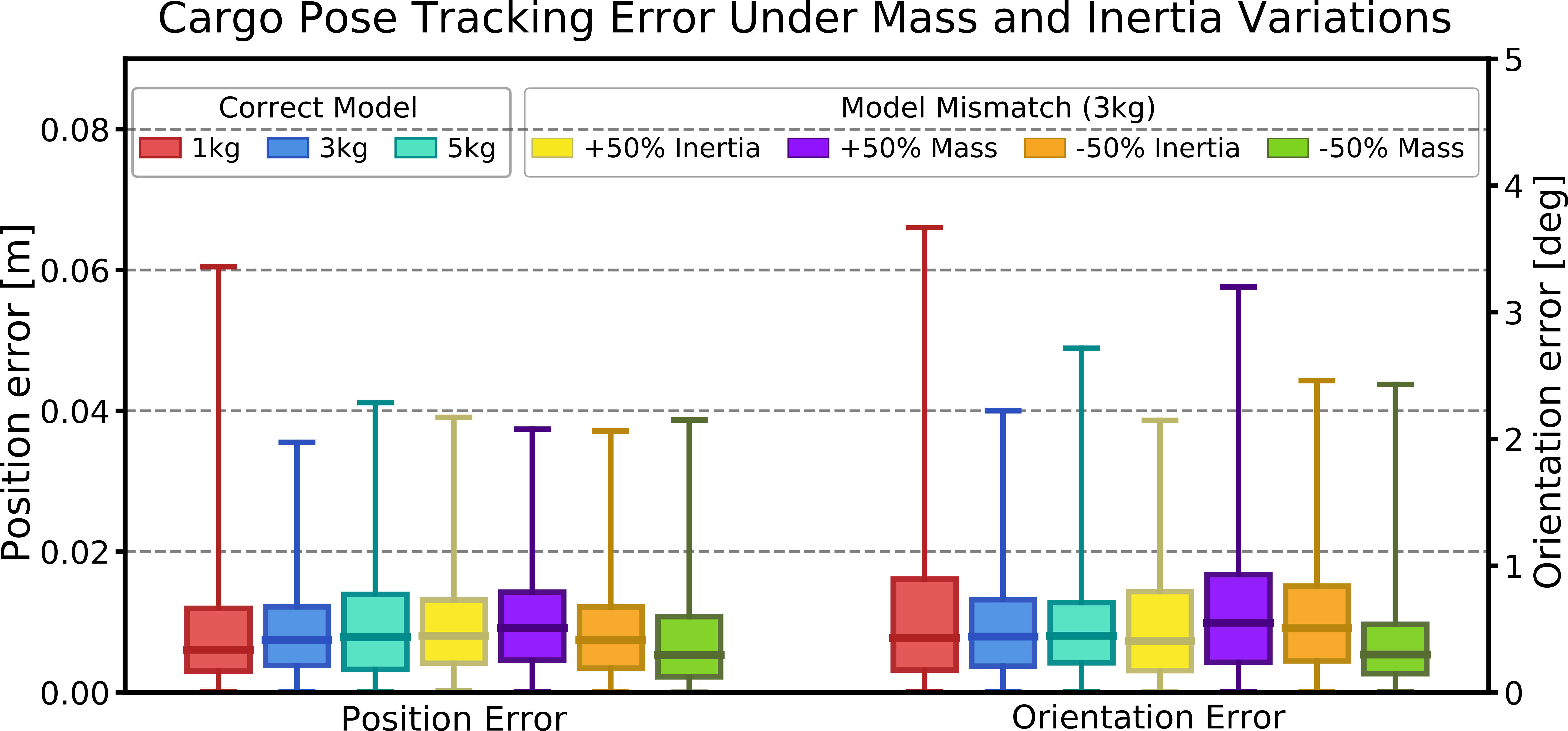}
    \vspace{-2.0mm}
    \caption{Cargo pose tracking errors under mass and inertia variations. 
    }
    \label{fig:robustness_analysis}
    \vspace{-2.0mm}
\end{figure}

To evaluate robustness, the robot--cargo system is commanded to track a reference trajectory consisting of a $6.5\,\mathrm{m}$ translation with a sharp $90^\circ$ turn in physical simulations. Multiple trials are performed under different nominal cargo masses as well as under deliberate mass and inertia modeling errors. 
As shown in Fig.~\ref{fig:robustness_analysis},
despite rare transient peaks caused by foot--ground impacts, the median errors and interquartile ranges remain consistently low across different nominal masses and under both mass and inertia modeling error, demonstrating strong robustness of the proposed controller. The controller performance degrades to failure when modeling error is increased to approximately $67\%$. 

\begin{figure}[h]
    \centering
    \includegraphics[width=0.85\linewidth]{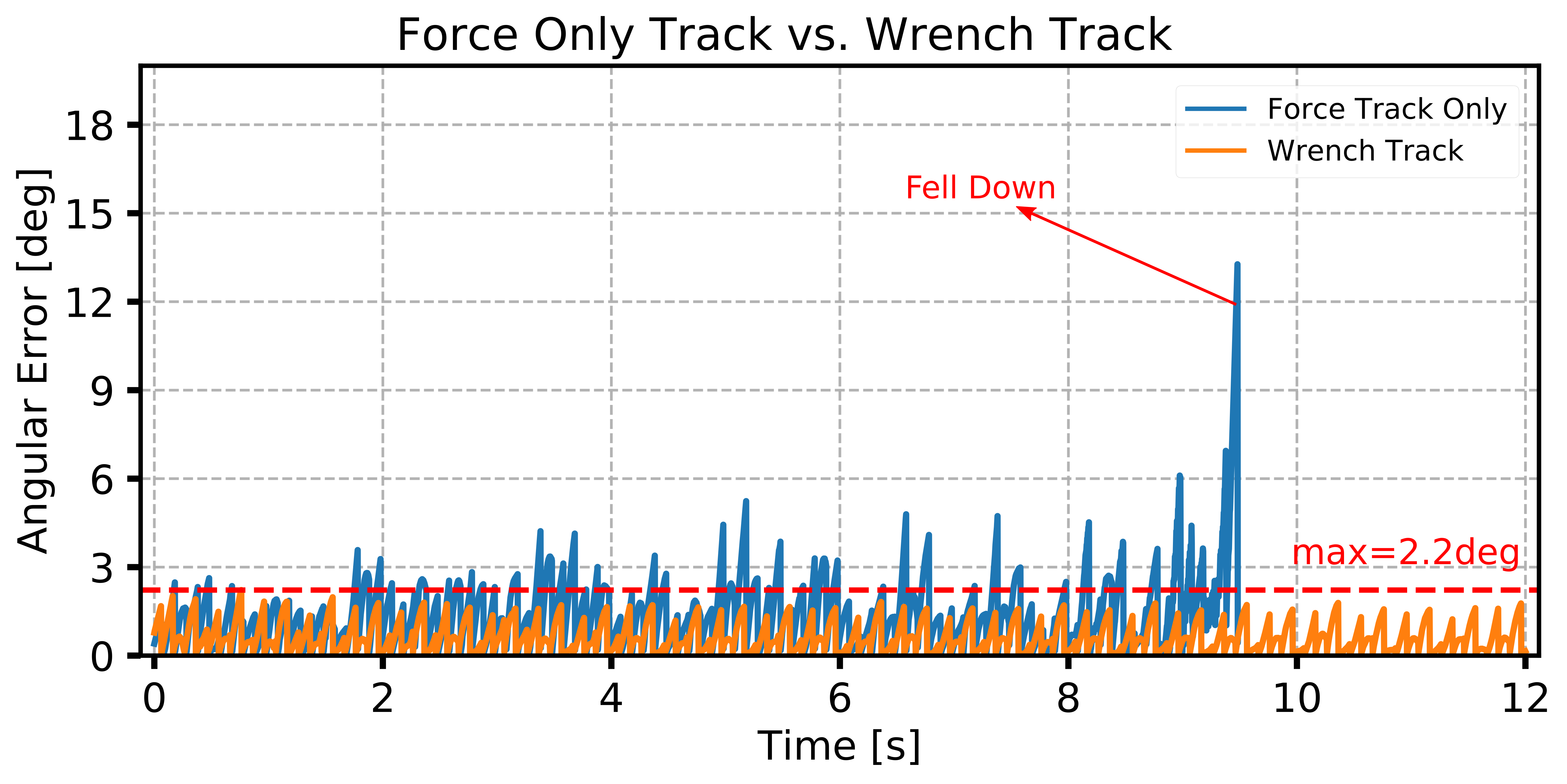}
    \vspace{-3.0mm}
    \caption{Comparison between force-only tracking and full wrench tracking.
    }
    \label{fig:wrench_ablation}
\end{figure}

Another important feature of our framework is that we optimize and track the full wrench, including both force and torque at the grasped handles. As an ablation study, when torque is completely disabled, the angular tracking error increases over time and eventually leads to instability, demonstrating the limitation of force-only tracking. Enabling torque tracking, particularly along the alignment axis between the robot and cargo (x-axis), significantly improves rotational stability and reduces angular error. The quantitative result is shown in Fig. \ref{fig:wrench_ablation}.






\section{Conclusion}
\label{sec:conclusion}
This paper presented an ADMM-based distributed MPC framework for collaborative prehensile loco-manipulation with multiple quadrupedal manipulators. Exploiting the star-shaped coupling induced by a shared payload, the centralized optimal control problem is decomposed into parallel robot-level subproblems, preserving dynamic coupling while improving scalability. Combined with a wrench-aware whole-body controller, the framework achieves real-time, force-consistent execution. Simulations with up to four robots validate scalability, fast convergence, and robustness to model and terrain variations. Future work includes hardware validation on physical platforms, and exploring GPU implementations to guide decentralized RL training.


\bibliographystyle{IEEEtran}
\bibliography{references}

\end{document}